\documentclass[conference]{IEEEtran}
\IEEEoverridecommandlockouts
\usepackage[utf8]{inputenc}
\usepackage{textcomp}
\usepackage{graphicx}
\usepackage[version=4]{mhchem}
\usepackage{siunitx}
\usepackage{longtable,tabularx}
\usepackage{accents}
\usepackage{comment}
\usepackage{mathtools}
\usepackage{subfigure}
\usepackage{caption}
\usepackage{color}
\usepackage{amsmath,amssymb}
\usepackage{booktabs}
\usepackage{tabu}
\usepackage{algorithm}
\usepackage{algpseudocode}
\usepackage{hyperref}
\usepackage{soul}
\usepackage[dvipsnames]{xcolor}
\usepackage{multirow}
\usepackage{enumerate}
\usepackage{svg}
\usepackage{epstopdf}

\usepackage{cite}

\newcommand{\od}[1]{{\color{green}OD: #1} }

\def\BibTeX{{\rm B\kern-.05em{\sc i\kern-.025em b}\kern-.08em
    T\kern-.1667em\lower.7ex\hbox{E}\kern-.125emX}}
\begin{document}

\title{Factor Graphs for Heterogeneous Bayesian Decentralized Data Fusion\\
%{\footnotesize \textsuperscript{*}Note: Sub-titles are not captured in Xplore and
%should not be used}
%\thanks{Identify applicable funding agency here. If none, delete this.}
}

\author{\IEEEauthorblockN{Ofer Dagan}
\IEEEauthorblockA{\textit{Smead Aerospace Engineering Sciences Dept.} \\
\textit{University of Colorado Boulder}\\
Boulder, USA \\
ofer.dagan@colorado.edu}
\and
\IEEEauthorblockN{Nisar R. Ahmed}
\IEEEauthorblockA{\textit{Smead Aerospace Engineering Sciences Dept.} \\
\textit{University of Colorado Boulder}\\
Boulder, USA \\
nisar.ahmed@colorado.edu}

}

\maketitle

\begin{abstract}
This paper explores the use of factor graphs as an inference and analysis tool for Bayesian peer-to-peer decentralized data fusion. 
We propose a framework by which agents can each use local factor graphs to represent relevant partitions of a complex global joint probability distribution, thus allowing them to avoid reasoning over the entirety of a more complex model and saving communication as well as computation cost.
%We suggest to opportunistically slice the factor graph into smaller, locally relevant, sub-graphs, representing ``chunks" of the global joint PDF, instead of reasoning over the full system graphical model. 
This allows heterogeneous multi-robot systems to cooperate on a variety of real world, task oriented missions, where scalability and modularity are key.   
%More specifically it involves heterogeneous multi-robot systems in real world, task oriented missions. We propose a shift in paradigm in the way decentralized data fusion is modeled and treated with probabilistic graphical models. Instead of reasoning over the full system graphical model we suggest to opportunistically slice the graph into smaller, locally relevant, sub-graphs, representing ``chunks" of the global joint PDF. 
%This is done by opportunistically breaking the problem into sets of local and common variables of interest, where the common set d-separates local  sets of variables. 
To develop the initial theory and analyze the limits of this approach, we focus our attention on static linear Gaussian systems in tree-structured networks and use Channel Filters (also represented by factor graphs) to explicitly track common information. We discuss how this representation can be used to describe various multi-robot applications and to design and analyze new heterogeneous data fusion algorithms. We validate our method in simulations of a multi-agent multi-target tracking and cooperative multi-agent mapping problems, and discuss the computation and communication gains of this approach.    
\end{abstract}

\begin{IEEEkeywords}
Bayesian decentralized data fusion (DDF), factor graphs, heterogeneous multi-robot systems, sensor fusion.
\end{IEEEkeywords}

\section{Introduction}
In the context of multi-agent systems, there is a growing interest to allow for the coordinated and cooperative operation of \emph{heterogeneous systems} in uncertain environments. For example a team of air and ground vehicles can vary in their 
sensing/computing/communication capabilities and/or might have a different set of models and objectives. The question is how to share information in a scalable and modular way, despite this heterogeneity?

\begin{figure}[bt]
      \centering
      %\framebox{\parbox{3in}{}
      \includegraphics[scale=0.49]{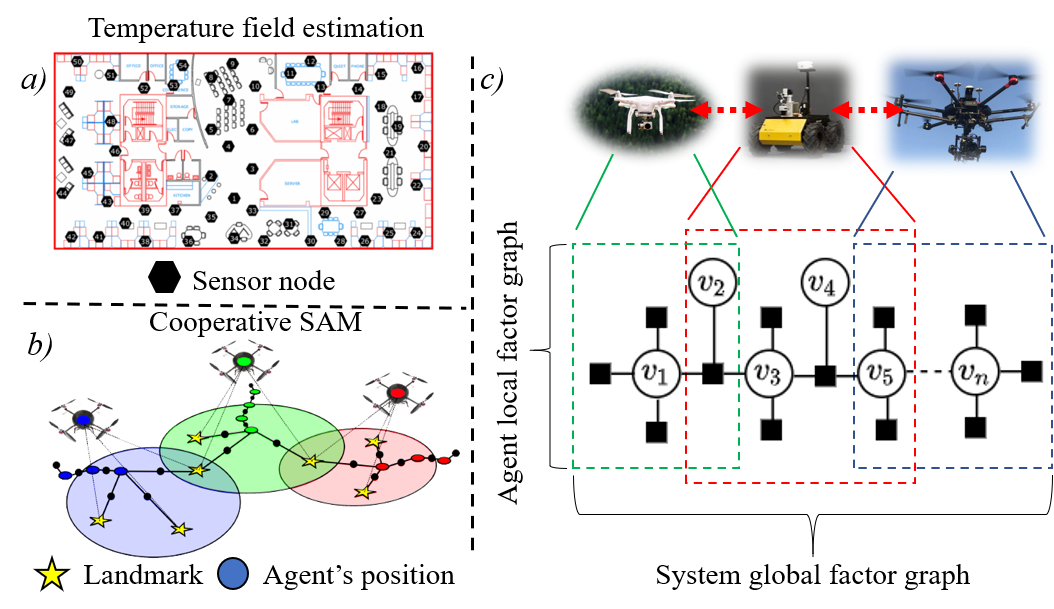}
      \caption{a) Application examples for cooperative heterogeneous tasks (figures adapted with permission from \cite{castello_temperature_2009}, \cite{cunningham_ddf-sam_2013}). b) A full system inference problem represented by a factor graph and its subdivision into agent's local graphs. Arrows show the tree-structured communication graph. }
      \label{fig:examples}
      \vspace{-0.22in}
\end{figure}

If we assume that the underlying task is to locally infer some quantity (state) of the system or environment, such as the temperature field or an environment map and individual robot positions in Fig. \ref{fig:examples}(a)-(b), respectively, then the problem translates to a data fusion problem. While there different methods for data fusion, this paper focuses on Bayesian decentralized data fusion (DDF). More specifically, this paper builds the foundations for DDF on factor graphs, to enable large scale multi-agent heterogeneous systems by reducing local communication and computation costs. The main idea is to allow each agent to only reason about its local task and communicate only the information relevant to its neighbor.

%In the context of multi-agent systems, \emph{heterogeneous systems} can refer to: different kinds of \emph{platforms}, such as air and ground vehicle cooperating on a joint task; different \emph{hardware and software} i.e., varied sensing/computing/communication capabilities; as well as different or overlapping \emph{models and objectives}.
%When these heterogeneous systems share information with the goal of inferring the probability density function (pdf) describing variables of interest, we define it as heterogeneous fusion.

For example, when cooperatively estimating the local biases and temperature field in a lab \cite{paskin_robust_2004} (Fig. \ref{fig:examples}(a)), it is clear that effect of far away sensors is negligible, while the influence of neighborhood sensors can't be ignored. Thus allowing each agent to process only locally relevant states instead of the full field might be crucial for scalability. Similarly, in the smoothing and mapping (SAM) application in \cite{cunningham_ddf-sam_2013} (Fig. \ref{fig:examples}(b)), allowing each agent to only reason about relevant parts of the map and its own position scales with the size of the map and network. But in both of these examples, for a correct estimate, we must properly account for the hidden correlations induced by the local states.

Proper treatment of the problem requires revisiting DDF theory, as it builds on the assumptions that all agents have the same monolithic/homogeneous model. 
%as notions of consistency and conservativity in the agent level are not well defined for heterogeneous problems and for general pdfs, albeit some progress is made in \cite{lubold_formal_2019}.
Further, identifying the conditional independence structure of the problem, to enable heterogeneous fusion, might not be trivial for other real world applications and complex dependencies. This raises the question, \emph{What is an appropriate representation for real world, mission and task oriented heterogeneous multi-robot systems, to allow for the design and analysis of decentralized data fusion algorithms?}

Probabilistic graphical models (PGMs) are a powerful and useful tool for inference problems \cite{pearl_probabilistic_2014}, where it is assumed that the full system graph, representing the joint distribution over all variables, is known and can be reasoned about. However, in decentralized systems this requires each agent to hold a copy of the full graph, irrespective to its local task or variables/states of interest. Instead, as shown in Fig. \ref{fig:examples}(c), we suggest ``slicing" the graph into local sub-graphs, such that each agent only reasons about the part of the graph relevant to its task.
  
%is usually not the case, as any agent might hold only a part of the graph, known locally. %This raises the question, \emph{how can it be leveraged to analyze and gain insight into heterogeneous decentralized data fusion (DDF) problems?}

The goal of this paper is to develop the theory and analyze the limitations of this approach using factor graphs. We show that factor graphs are a natural choice for this aim, as they can  represent a variety of robotics applications with the focus on heterogeneous DDF problems. We present \emph{FG-DDF} - a formulation of the DDF problem in terms of factor graphs. We set the foundations by formulating the problem and defining graph operations in general terms when possible. We show that the fusion operation becomes almost trivial, as messages between agents can be treated as ``factors" simply added to the local graph at the receiving agent. Further we show how common information can be identified (when it can be defined), especially in heterogeneous fusion problems. To explicitly demonstrate the algorithms presented, we focus on static linear Gaussian problems in tree-structured networks and leave conservative marginalization of past states in dynamic problems and ad-hoc network topologies to future work. Finally, we present simulation results which show significant computational ($95\%$) and communications ($97\%$) savings compared to homogeneous fusion.

\section{Problem Statement}
\label{sec:ProbStatement}
%\subsection{Problem Statement}
Let $V$ be a global set of random variables describing  states of interest monitored by a set $N_a$ of $n_a$ autonomous agents. The states of interest are distributed between the agents such that each agent $i$ monitors some ``local states of interest", which are a subset of the global set $\chi^i\subseteq V$. The set $\chi^i$ can be further divided into a set of local states $\chi^i_L$, observed locally by agent $i$, and a set of common states between agent $i$ and its neighbors $\chi^i_C=\bigcup_{j\in N_a^i}^{}\chi^{ij}_C$, where $N^i_{a}\subseteq N_a$ is the set of agents communicating with agent $i$ and $\chi^i=\chi^i_L\bigcup \chi^i_C$.

Each agent is able to gather information about its local states of interest, $\chi^i$, by (i) taking independent local measurements $Y_k^i$, described by the measurement likelihood $p(Y^i_{k}|\chi^i)$ and (ii) by fusing messages from a neighboring agent $j\in \{1,...,N_a, i\neq j\}$ regarding common states $p^j(\chi^{ij}_C|Z^j_k)$, where $\chi^{ij}_C\subseteq \chi^{i}_C$ are the common states to agents $i$ and $j$ and $Z^j_k$ is all the information available at agent $j$ up to and including time step $k$. In the above definitions $\chi$ can represent a static or dynamic state, and we neglect states time indexing for ease of notation.

By allowing the pdfs to represent distributions over identical or overlapping sets of states, along with the flexible definition of the local and common sets of states, we enable the analysis of a variety of robotics and data fusion applications, such as static/dynamic, homogeneous/heterogeneous problems, under one framework. For instance, in the temperature sensing static example of Fig. \ref{fig:examples}(a), sensor $i$ bias state would belong to the local set $\chi^i_L$ and a neighboring sensor $j$ temperature measurement would belong to the common set $\chi_C^{ij}$. Similarly, in the dynamic SAM example of Fig. \ref{fig:examples}(b), agent $i$ position states would be $\chi^i_L$ and landmarks communicated from agent $j$ in $\chi^{ij}_C$.

%if we set $\chi^i_L=\varnothing$ and $\chi^{ij}_C=\chi,\ \forall i,j\in N_a$ the fusion equation (\ref{eq:probStatement}) defines homogeneous fusion, since then $\chi^i=\chi^j=\chi$. But if we set $\chi^i_L\neq \varnothing$ and $\chi^{ij}_C\neq \chi$ then $\chi^i\neq \chi^j$ and eq. (\ref{eq:probStatement}) defines heterogeneous fusion.

The DDF problem is defined as locally inferring the posterior distribution based on all available information at agents $i$ and $j$ up to and including time step $k$. Here we look for a fusion function $\mathbb{F}$, that takes as input the agents prior distributions $p^i(\chi^{i}|Z^{i,-}_{k})$ and $p^j(\chi^{j}|Z^{j,-}_{k})$ and returns the locally fused posterior distributions $p^i_f(\chi^{i}|Z^{i,+}_{k})$ and $p^j_f(\chi^{j}|Z^{j,+}_{k})$:
\begin{equation}
    \begin{split}
        (p^i_f(\chi^{i}|Z^{i,+}_{k}), p^j_f(\chi^{j}|Z^{j,+}_{k})) = \mathbb{F}(p^i(\chi^{i}|Z^{i,-}_{k}), p^j(\chi^{j}|Z^{j,-}_{k})),
    \end{split}
    \label{eq:probStatement}
\end{equation}
where $Z^{j,-}_k \equiv Y^j_{k}\cup Z^j_{k-1}$ is the local information at agent $j$ before fusion at time step $k$, and $Z^{i,+}_k$, $Z^{j,+}_k$ is local information at the agents after fusion. Note that in general $Z^{i,+}_k\neq Z^{j,+}_k$ as in the case of heterogeneous fusion, where information regarding local states is not shared.

In this paper we suggest solving the decentralized inference problem with factor graphs, where each agent holds a factor graph that represents the local posterior distribution, which is the result of a fusion rule given in  (\ref{eq:probStatement}). Fusion of new information, whether it is the result of local observations ($Y^i_k$) or of information from a neighboring agent ($Z_k^j$) is performed by adding a factor to the local graph, and estimation of the states of interest is done by local inference on the graph. We claim that this is a shift in paradigm for how to analyze and design DDF fusion algorithms, where instead of maintaining a graph over the full system variables $V$, each agent holds a smaller graph, only over the subset $\chi^i$.

\subsection{Related Work}
Algorithms for heterogeneous fusion, as detailed in \cite{dagan_exact_2021}, are advantageous as they scale with a subset of locally relevant variables instead of the number of agents in the system. However this is not without cost, as it requires approximations in order to remove indirect dependencies between variables not mutually monitored by both agents. The problem becomes acute in filtering scenarios, where marginalization results in hidden correlations.  In \cite{dagan_exact_2021} a family of heterogeneous fusion algorithms, conditionally factorized channel filters (\emph{CF$^2$}), is developed by utilizing the conditional independence structure in a multi-agent multi-target tracking application. It is shown that even in this limited setting of \cite{dagan_exact_2021} (tree-structured network and full communication rate), challenges arise in identifying the common information and in properly accounting for it, as in the case of fixed-lag smoothing or filtering problems. 

In \cite{paskin_robust_2004} Paskin \textit{et al.} present a distributed inference problem in a network of static sensors. It is solved by using a robust message passing algorithm on a junction tree, but this is limited to static variables and requires the full construction of the tree before performing inference. In \cite{makarenko_decentralised_2009} Makarenko \textit{et al.} extend Paskin's algorithm for a dynamic state and formulate it as a DDF problem, however, the algorithm is limited to a single common state, i.e., the focus is on a homogeneous problem. Assuming static network topology for a static model they show equivalence between the channel filter (CF) \cite{grime_data_1994} and the Hugin \cite{andersen_hugin_1989} algorithms. 
In \cite{chong_graphical_2004} Chong and Mori use both Bayesian networks and information graphs to identify conditional independence and track common information, respectively. While they use conditional independence to reduce state dimension, it is only for communication purposes and not for local state reduction. Further, their graphical model representation is used for analysis purposes and not for inference on the graph itself.

Thus, PGMs for decentralized data fusion is not yet utilized to the full extent: for both analyzing problem structure by identifying conditional independence and for efficient inference on the graph. 
Here we suggest using factor graphs to leverage the sparsity induced by the decentralized structure of the problem.

Factor graphs are used in a variety of robotic applications \cite{dellaert_factor_2021} and are beneficial in (i) providing insight into the structure of problems across robotics, (ii) enabling efficient inference algorithms and (iii) analysing the correctness of local inference. Despite these benefits, to the best of our knowledge, previous work only looked at limited/niche cases of heterogeneous fusion \cite{cunningham_ddf-sam_2013}, or when agents are considered as neighbors only if they share a relative measurement \cite{etzlinger_cooperative_2017}, so present work endeavors for a wider and more general framework.

%\vspace{-0.1in}

\section{Bayesian Decentralized Data Fusion}
\label{DDF}
Given two agents i and j, we are interested in forming an estimate of the state $\chi$ given all the information available locally, defined by $Z^i$ and $Z^j$, respectively. Assuming all measurements are conditionally independent given the state $\chi$ and using a distributed variant of Bayes' rule, \cite{chong_distributed_1983} shows that the exact state estimate can be achieved using:
\begin{equation}
    p_f(\chi|Z^{i,-}_k\cup Z^{j,-}_k)\propto \frac{p^i(\chi|Z^{i,-}_k)p^j(\chi|Z^{j,-}_k)}{p^{ij}_c(\chi|Z^{i,-}_k\cap Z^{j,-}_k)}
    \label{eq:Homogeneous_fusion}
\end{equation}
where $p^{ij}_c(\chi|Z^{i,-}_k\cap Z^{j,-}_k)$ is the distribution of the common information (when it makes sense to define one) of the two agents which needs to be removed in order to avoid double counting. Notice that the above fusion rule satisfies  (\ref{eq:probStatement}) for the homogeneous case $\chi=\chi^i=\chi^j$. \\
In this paper we assume that common information is explicitly tracked, by either keeping a pedigree of all information transformed in fusion instances or by restricting network topology to a tree structure, as in the channel filter \cite{grime_data_1994}.

For heterogeneous fusion, $\chi^i\neq \chi^j$, assuming the local states are conditionally independent given common states:
\begin{equation*}
    \chi^i_L\perp \chi^j_L \mid \chi^{ij}_C,
\end{equation*}
Dagan \textit{et al.} show in \cite{dagan_exact_2021} that agents can fuse only a subset of common variables without the need to communicate ``irrelevant" local variables. The fusion rule at agent $i$ can be written as,
\begin{equation}
    \begin{split}
        p^i_f(\chi^i|Z^{i,+}_k)\propto
        \frac{p^i(\chi^{ij}_C|Z^{i,-}_k)p^j(\chi^{ij}_C|Z^{j,-}_k)}{p^{ij}_c(\chi^{ij}_C|Z^{i,-}_k\cap Z^{j,-}_k)}\cdot p^i(\chi^i_L|\chi^{ij}_C,Z^{i,-}_k).
    \end{split}
    \label{eq:Heterogeneous_fusion}
\end{equation}
In \cite{dagan_exact_2021}, assuming tree topology, this is named the \emph{heterogeneous-state channel filter} (HS-CF).
As we show later, (\ref{eq:Heterogeneous_fusion}) amounts to sending/incorporating a factor in the local graph of each of the agents $i/j$.

We move our attention to factor graphs, we describe and define them for general pdfs and for the special case of Gaussian distributions (Sec. \ref{sec:FactorGraphs}). We then put it all together in Sec. \ref{sec:FG_DDF} to formalize factor graphs for DDF.   

\section{Factor Graphs}
\label{sec:FactorGraphs}
\begin{figure}[b]
	\centering
	\includegraphics[width=0.52\textwidth]{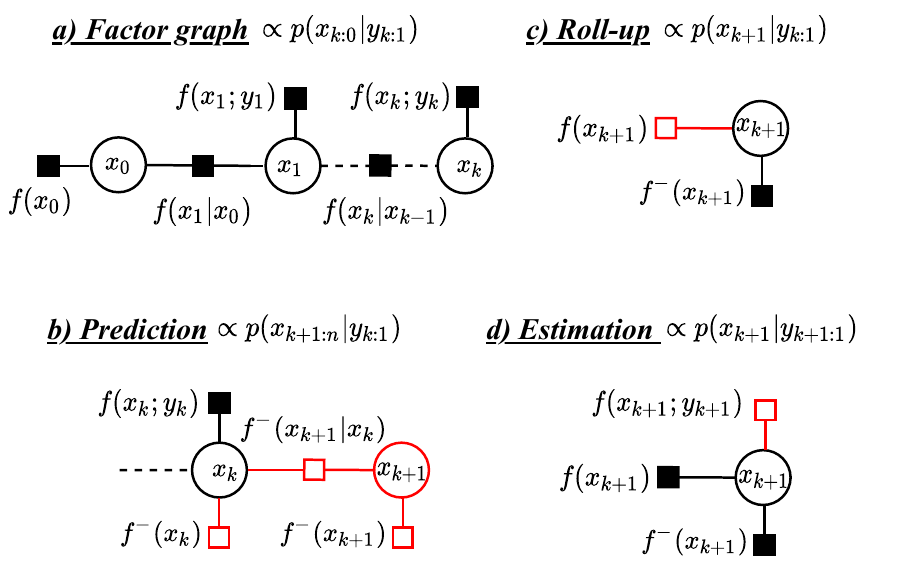}
	\caption{a) Example factor graph representing a tracking scenario, where the variables of interest are the target states from time $t=0$ to $t=k$. Shown are the different types of factors (prior, transition and measurement). b-d) show the graph operations, namely prediction, roll-up and estimation, with new factors and variables marked by red empty nodes.  } 
	\label{fig:factorGraph_operations}
	\vspace{-0.1in}
\end{figure}

A factor graph is an undirected bipartite graph $\mathcal{F}=(U,V,E)$ with factor nodes $f_l\in U$, random variable nodes $v_m\in V$ and edges $e_{lm}\in E$ connecting factor node \emph{l} to variable node \emph{m}. Each factor $f_l$ has one set of neighboring variable nodes $V_l\in V$ such that $f_l(V_l)$ is a function of only those variables in $V_l$. The joint distribution over the graph is then proportional to the global function $f(V)$, factorized by $f_l(V_l)$:
\begin{equation}
    p(V)\propto f(V)=\prod_{l}f_l(V_l)
    \label{eq:factorization}
\end{equation}

For a more complete description of factor graphs and algorithms for different operations on the graphs see \cite{frey_factor_1997}. For factor graphs in the context of robotics and graph-SLAM see \cite{dellaert_factor_2017}.

\begin{comment}

\begin{figure}[tb]
      \centering
      %\framebox{\parbox{3in}{}
      \missingfigure[figwidth=6cm]{Put an example factor graph here}
      \caption{}
      \label{fig:factorGraph}
      %\vspace{-0.2in}
\end{figure}
\end{comment}
An example of a factor graph describing a simple tracking problem, where the variables $v_m$ describe the target position states in time $x_k$, can be seen in Fig. \ref{fig:factorGraph_operations}a. Here 
we have three types of factors: a prior, a measurement likelihood and a dynamic transition factors. The posterior distribution, given the measurements, is then proportional to the factorization of the global function:
\begin{equation}
    \begin{split}
        p(x_{k:0}|y_{k:1})\propto f(x_0)\prod_{k=1}^{K}f(x_k;y_k)f(x_k|x_{k-1}),
    \end{split}
\end{equation}
where we use $f(x_k|x_{k-1})=f(x_k,x_{k-1})\propto p(x_k|x_{k-1})$ and $f(x_k;y_k)=f(x_k)\propto l(x_k;y_k)$ to express and emphasize conditional probabilities between variables and between variables to measurements, respectively. This is based on the idea in \cite{frey_extending_2002} of extending factor graphs to represent directed (Bayesian networks) and undirected (Markov random fields), thus making dependencies explicit when reading the factor graph.\\   
As we show in more detail later, in DDF, fusion is done by incorporating a new factor into the local graph, representing the new information that agent $i$ received from agent $j$. 

\subsection{Factor Graphs in Gaussian Canonical Form}
A factor graph $F=(U,V,E)$ represents the distribution $p(V)$ by a product of its factors (see (\ref{eq:factorization})). This can be transformed to a summation by converting the factors to log-space as described in \cite{koller_probabilistic_2009}. A special case is the canonical (information) form of linear Gaussian distributions; factors then consists of two elements, namely the information vector and matrix, that in turn can be summed over the graph to describe the multivariate Gaussian distribution in canonical form,
\begin{equation}
    p(V)\sim \mathcal{N}(\zeta, \Lambda)= \sum_{l}f_l(V_l).
    \label{eq:canonicalFactorization}
\end{equation}
Here $\zeta$ and $\Lambda$ are the information vector and matrix, respectively. In this case all factors are Gaussian distributions represented in canonical form $f_l(V_l)\sim \mathcal{N}(\zeta_l, \Lambda_l)$.

\subsection{Probabilistic Operations on Factor Graphs}
In \cite{paskin_thin_2002} a graphical model perspective of the SLAM problem is discussed. In order to reduce the size of the graph a filtering approach is taken by defining three types of operations on the dynamic Bayes net (DBN) type graph. The three operations are: \emph{prediction}, \emph{roll-up} (marginalization of past states) and \emph{estimation} (measurement update). We follow \cite{paskin_thin_2002} and the information augmented state (\emph{iAS}) filter presented in \cite{dagan_exact_2021} to define these operations on a factor graph and show their translation into new factors.\\
\emph{Prediction:}\\
In the prediction step three factors are added to the graph: two unary factors $f^-(x_k), f^-(x_{k+1})$, connected to the variable nodes $x_k$ and $x_{k+1}$, respectively, and a binary factor $f^-(x_{k+1}|x_k)$ connected to both variables and describes the correlation between the two variables. The $(-)$ superscript denotes prediction. Each factor is defined by an information vector and matrix $\{\zeta, \Lambda\}$:
\begin{equation}
    \begin{split}
        f^-(x_k) &= \{-F_k^TQ_k^{-1}G_ku_k,\ F_k^TQ_k^{-1}F_k\} \\
        f^-(x_{k+1}) &= \{Q_k^{-1}G_ku_k, \ Q_k^{-1} \} \\ 
        f^-(x_{k+1}|x_k) &= \left \{ \begin{pmatrix}
        0_{n\times1} \\ 0_{n\times1}
        \end{pmatrix}, \begin{pmatrix}
        0_{n\times n} & -Q_k^{-1}F_k\\ 
        -F_k^TQ_k^{-1} & 0_{n\times n}
        \end{pmatrix} \right \}.
    \end{split}
    \label{eq:predFactors}
\end{equation}
Here $F_k$ and $G_k$ are the state transition and control matrices at time step $k$, respectively. $u_k$ is the input vector and $Q_k$ is a zero mean white Gaussian process noise covariance matrix. The graphical description of the prediction step is given in Fig. \ref{fig:factorGraph_operations}b.\\
\emph{Roll-up (marginalization):}\\
It is known that upon marginalization of a random variable $x$ in a directed graphical model, all variables in the Markov blanket of $x$ are moralized, i.e. ``married'' by adding an edge. The effect in a factor graph is similar, and variables are moralized by adding a factor connecting all variables in the Markov blanket of the marginalized variable. If we denote the Markov blanket of $x$ by $\bar{x}$, then the new factor $f(\bar{x})$ is computed in two steps:\\
1. Sum all factors connected to $x$ to compute the following information vector and matrix:
\begin{equation}
    \{\zeta,\ \Lambda \} = f(x)+\sum_{i\in \bar{x}}f(x,\bar{x}_i)
\end{equation}
2. Use Schur complement to compute the marginal and find the new factor $f(\bar{x})$:
\begin{equation}
    f(\bar{x}) = \{\zeta_{\bar{x}}-\Lambda_{\bar{x}x}\Lambda_{xx}^{-1}\zeta_x,\  \Lambda_{\bar{x}\bar{x}}-\Lambda_{\bar{x}x}\Lambda_{xx}^{-1}\Lambda_{x\bar{x}}\}.
    \label{eq:marginalFactor}
\end{equation}
Notice that as a result, conditionally independent variables become correlated.\\
We demonstrate marginalization in Fig. \ref{fig:factorGraph_operations}c, marginalizing out $x_{k:n}$ induces a new factor $f(x_{k+1})$ over $x_{k+1}$. Here, since the only variable in the Markov blanket of the marginalized variables is $\bar{x}=x_{k+1}$, the new factor is unary over $x_{k+1}$ alone.\\
\emph{Estimation (measurement update):}\\
Adding a measurement in the canonical (information) form of the Kalman filter is a simple task. In the factor graph this translates to adding a factor $f(x_{k+1};y_{k+1})$ connected to all measured variables,
\begin{equation}
    f(x_{k+1};y_{k+1}) = \{H_{k+1}^TR_{k+1}^{-1}y_{k+1},\ H_{k+1}^TR_{k+1}^{-1}H_{k+1} \}.
    \label{eq:measFactors}
\end{equation}
Where $H_{k+1}$ is the sensing matrix, $R_{k+1}$ is a zero mean white Gaussian measurement noise covariance matrix and $y_{k+1}$ is the noisy measurement vector.
Figure \ref{fig:factorGraph_operations}d shows the addition of a unary measurement factor $f(x_{k+1};y_{k+1})$.
\begin{comment}

\begin{figure}[tb]
      \centering
      %\framebox{\parbox{3in}{}
      \missingfigure[figwidth=6cm]{Put an example of the graph operations here}
      \caption{}
      \label{fig:factorGraph_operations}
      %\vspace{-0.2in}
\end{figure}
\end{comment}

\section{Factor Graphs for DDF - FG-DDF}
\label{sec:FG_DDF} 
The shift in our approach to PGM based decentralized data fusion is that now each agent holds its own \emph{local} factor graph (Fig. \ref{fig:fusion}a and \ref{fig:fusion}b) and does not need to reason over a full (global) version of the graph. The fusion problem can be viewed as exchanging factors between agents, and thus the challenge is to account for common information, i.e. guarantee that mutual information is not introduced more than once into the local graph. While this is a common problem in DDF, the advantage of the graph approach, especially in the case of heterogeneous fusion, is in utilizing the structure evident from the graph to account for the common information to design and analyze fusion algorithms. We now show how the fusion equations translate to a factor graph representation and then demonstrate its applicability in analyzing the HS-CF algorithm \cite{dagan_exact_2021}.  

\subsection{DDF in Factor Graphs} 
\begin{figure*}[bt]
	\centering
	\includegraphics[width=1\textwidth]{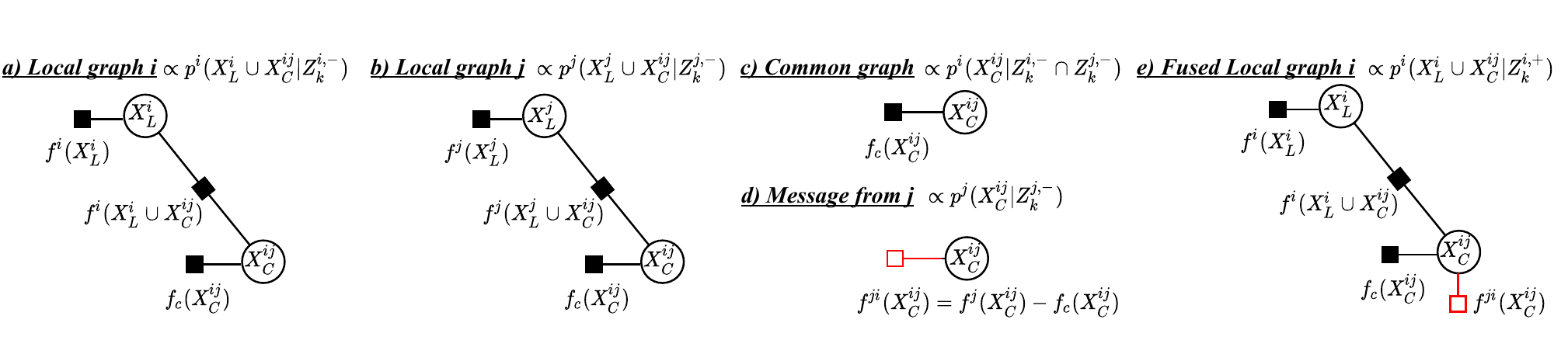}
	\caption{DDF in factor graphs: a and b showing the local factor graphs over each agent's variables of interest. c) is the common graph describing the factors common to both agents. d-e) demonstrates the fusion operation; the message sent from $j$ with new information over the common variables (d) then integrated into $i$'s local graph with a simple factor addition (e).} 
	\label{fig:fusion}
	\vspace{-0.2in}
\end{figure*}

Taking natural logarithm and rearranging, the heterogeneous fusion given in (\ref{eq:Heterogeneous_fusion}) can be written as,
\begin{equation}
    \begin{split}
        & \log p^i_f(\chi^i|Z^{i,+}_k) = \log p^i(\chi^i_L|\chi^{ij}_C,Z^{i,-}_k)+\log p^i(\chi^{ij}_C|Z^{i,-}_k)\\
        &+\log p^j(\chi^{ij}_C|Z^{j,-}_k) 
        -\log p_c(\chi^{ij}_C|Z^{i,-}_k\cap Z^{j,-}_k)+\Tilde{C}
    \end{split}
    \label{eq:Log_Heterogeneous_fusion}
\end{equation}
We can divide the terms in the above equation into two contributions; the first row corresponds to the information agent $i$ holds prior to fusion while the second row (ignoring the normalization constant) describes the new information agent $i$ should receive from agent $j$ regarding their common states $X_C^{ij}$, assuming the common information is explicitly known. This message can be viewed as sending a factor $f^{ji}(X_C^{ij})$ from agent $j$ to agent $i$, which is added into agent $i$'s factor graph as seen in Fig. \ref{fig:fusion}d and \ref{fig:fusion}e.

The pseudo code for the FG-DDF Algorithm is given in Algorithm \ref{algo:FG-DDF}. This represents the local fusion algorithm for each each agent $i$ communicating with an agent $j$ in its neighborhood $N_a^i$. In the case of tree structured network, if the agents are using a channel filter to track the common information, then an additional factor graph is instantiated over the common states for each communication channel agent $i$ has with its neighbors (line 5). The main steps of the algorithm are the send / fuse message (lines 13-14), since these steps are specific to each fusion algorithm, we choose the HS-CF as an example and give pseudo code in Algorithms \ref{algo:sendMsg} and \ref{algo:fuseMsg}, respectively.

\begin{algorithm}[bt]
    \caption{FG-DDF}
    \label{algo:FG-DDF}
    \begin{algorithmic}[1]
    \State Define: $\chi_{i}$, Priors, Fusion algorithm %\Comment{{\color{ForestGreen} Fig. }}
    \State Initialize local factor graph $\mathcal{F}^{i}$
    \If{CF algorithm}\For{all $j \in N_a^i$}
    \State Initialize CF factor graph $\mathcal{F}^{ij}$ over $\chi_C^{ij}$
    \EndFor
    \EndIf 
    \For{all time steps}
    \For{all $x \in \chi_{i}$}
    \State Measurement update step
    \Comment{{\color{Gray} Eq.\ref{eq:measFactors} }}
    \EndFor
    \For{all $j \in N_a^i$}
    \State Send message to $j$ \Comment{{\color{Gray} e.g., Algorithm \ref{algo:sendMsg} }} 
    \State Fuse message from $j$
     \Comment{{\color{Gray} e.g., Algorithm \ref{algo:fuseMsg} }}   
    \EndFor
    \EndFor
    \State \Return
    \end{algorithmic}
\end{algorithm}

\subsection{Inference}
In the previous sections we described how an agent builds a local factor graph over its variables of interest through a series of predictions, local measurements and fusion of data from neighboring agents. At this point it is important to describe how, given the local factor graph, inference is performed. The goal of inference in this case is to deduce the marginal pdfs of each variable of interest from the joint function described by the factor graph. When the factor graph is tree-structured the \emph{sum-product algorithm} \cite{frey_factor_1997}, \cite{kschischang_factor_2001} can be used to directly work on the factor graph. However, when there are more than one variable in common between two communicating agents, the post-fusion graph has cycles due to the marginalization of the local variables at the ``sending" agent $j$.

\begin{figure}[bt]
	\centering
	\includegraphics[width=0.45\textwidth]{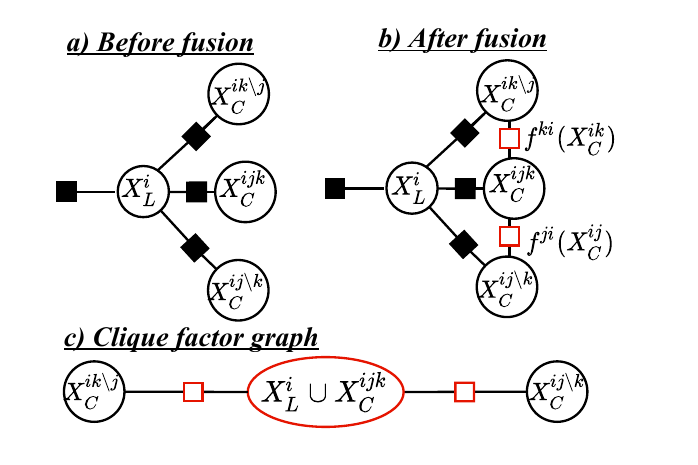}
	\caption{Transitioning the graph into a clique factor graph demonstrated on a network $j-i-k$. a) Local graph before fusion - tree structure. b) Messages sent from $k$ and $j$ to $i$ results in loops in the local graph. c) a tree structure is regained by forming cliques.  } 
	\label{fig:clique_fg}
	\vspace{-0.1in}
\end{figure}

To solve this problem we suggest transforming the factor graph into a tree by forming cliques over the cycles. Note that we are not transforming the graph into a clique or a junction tree, but only forming the minimum number of cliques that will result a tree graph as we explain next. It is worth mentioning here that Kaess \textit{et al.} \cite{kaess_bayes_2011} use the elimination algorithm to transfer a factor graph into a Bayes tree for incremental updates. This approach might be useful in the future but is not advantageous in the scope of this work. Instead, we keep the graph as a factor graph and only join into cliques variables which are now not d-separated by the ``local" variables and then summarize their factors to new joint factors, connected to the clique. This is demonstrated in Fig. \ref{fig:clique_fg}, where assuming a tree-structured network $j-i-k$, the common variables are separated into three different sets $X_C^{ijk}$, $X_C^{ij\setminus k}$ and $X_C^{ik\setminus j}$, representing variables common to the three agents, variables common to $i$ and $j$ but not to $k$ and similarly variables common to $i$ and $k$ but not to $j$, respectively. We can see that fusion results loops in the graph. We restore the tree structure by forming a clique over the separation sets $X_L^i \cup X_C^{ijk}$. Note that this needs to be done only before inference and not for fusion or filtering.  

\subsection{HS-CF in Factor Graphs}
To demonstrate the applicability in analyzing algorithms and track the implicit common information in heterogeneous DDF problems we show how the heterogeneous state channel filter (HS-CF) algorithm \cite{dagan_exact_2021} translates to a factor graph.
Briefly, the HS-CF builds on the homogeneous channel filter (CF) \cite{grime_data_1994} and exploits conditional independence between states local to each agent, given the common states between the agents as described in (\ref{eq:Heterogeneous_fusion}), to perform heterogeneous fusion and only fuse information relating to common states of interest. When the network is tree-structured, a channel filter is placed on the communication channel between any two agents, tracking the information passed through the channel. In a factor graph this translates to each agent maintaining a \emph{``locally full"} factor graph over all its states of interest $\chi^i$ and a separate \emph{partial} factor graph over the common states of interest $\chi_C^{ij}$ for each communication channel, as shown in Fig. \ref{fig:fusion}a-c respectively.

Fusion over common states of interest is done in two steps, sending a message and fusing incoming messages. When sending a message, as described in Algorithm \ref{algo:sendMsg}, an agent first marginalizes out local ``irrelevant" variables out of its graph (Algorithm \ref{algo:marginalize}), then subtracts the channel filter factors to synthesize new information factors to send to agent $j$ and updates the CF graph with the new factors.

Fusing an incoming message is a simple process, as described in Algorithm \ref{algo:fuseMsg}. An agent $i$ that received a message from a neighbor agent $j$ need only to update its local graph and CF graph with the set of incoming factors over the common variables of interest.

\begin{algorithm}[bt]
    \caption{Send Message to $j$ (HS-CF)}
    \label{algo:sendMsg}
    \begin{algorithmic}[1]
    \State {Input: $j$ - agent to communicate with} 
    \State Marginalize out local variables $\chi^i\setminus  \chi_C^{ij} $  \Comment{{\color{Gray}  Algorithm \ref{algo:marginalize}}} 
    \State Subtract factors of CF factor graph from factors of marginalized graph to receive a set of new information factors $f\in \pmb{f}^{ij}(\chi_C^{ij})$
    \State Update CF factor graph with all $f_n\in \pmb{f}^{ij}(\chi_C^{ij})$
    \State \Return Message $\pmb{f}^{ij}(\chi_C^{ij})$
    \end{algorithmic}
   % \vspace{-0.1in}
\end{algorithm}

\begin{algorithm}[bt]
    \caption{Fuse Message from $j$ (HS-CF)}
    \label{algo:fuseMsg}
    \begin{algorithmic}[1]
    \State {Input: Message from agent $j$ - $\pmb{f}^{ji}(\chi_C^{ij}) $} 
    \For{all $f \in \pmb{f}^{ji}(\chi_C^{ij})$}
    \State Add $f$ to local factor graph $\mathcal{F}^{i}$
    \State Add $f$ to CF factor graph $\mathcal{F}^{ij}$
    \EndFor
    \State \Return
    \end{algorithmic}
\end{algorithm}

\begin{algorithm}[bt]
    \caption{Marginalize variable $x$ from a factor graph $\mathcal{F}$}
    \label{algo:marginalize}
    \begin{algorithmic}[1]
    \State {Input: variable to remove $x$, factor graph $\mathcal{F}$} 
    \State Sum $f(x)$ and all factors $f(x,\bar{x}_i)$ adjacent to $x$
    \State Create new marginal factor $f(\bar{x})$ using (\ref{eq:marginalFactor})
    \State Add edges from every $\bar{x}_i \in \bar{x}$ to $f(\bar{x})$
    \State Remove $f(x,\bar{x}_i)$ factor and $x$ from $\mathcal{F}$
    \State \Return marginal factor graph $\mathcal{F}$
    \end{algorithmic}
\end{algorithm}

%\section{Application Examples}
%\label{FactorAlgo}
%\input{Text/FactorAlgorithms}
%\subsection{Paskin and Guestrin Tmp+bias estimation}
%\subsection{Cunningham - DDF-SAM}

\section{Simulation}
\label{sec:Sim}
To validate the algorithms and demonstrate our approach for decentralized fusion problems we perform two different simulations. First we validate the algorithms with a multi-agent multi-target tracking simulation and compare it to previous simulation results, which were implemented using conventional KF-based filtering and HS-CF equations as described in \cite{dagan_exact_2021}. Then we test the algorithm on a multi-agent mapping problem and show its applicability and advantages in computation and communication reduction, thus making it scalable and attractive for large networks. 
The algorithms are written in Python and are based on the \emph{fglib} \cite{bartel_danbarfglib_2021} and \emph{NetworkX} \cite{hagberg_exploring_2008} open source libraries

\subsection{Example 1 - Target tracking}
As a validation case, we repeat and compare to the simulation scenario described in \cite{dagan_exact_2021}, and so for brevity, we will not repeat all the details here. We consider a 2D static five-agent six-target tracking problem, where the agents are connected bi-laterally in ascending order. Each agent $i$ inference task is to estimate it own constant relative position measurement bias (assuming perfect local position information) and the position of a set of $n_t^i$ targets. 
The agent's target assignments and sensor measurement error covariances are given in Table \ref{tab:measurementError}.
In the PGM heterogeneous formulation of the problem, each agent $i$ maintains a local factor graph over $\chi^i=\cup_{t\in T^i}x_t^i\bigcup s^i$, where $x_t^i=[e_t,n_t]^T$ are the east and north positions of target $t$ in its target assignments $T^i$, and $s^i=[b_{e}^i,b_{n}^i]^T$ is the agent's own east and north biases. 
In the homogeneous formulation, each agent has to process and communicate the full set of variables (22 in this example). In heterogeneous fusion, it processes and communicates only a subset of the full system variables: maximum 8 and 4, respectively (e.g., agent 3). This leads to about $2.6\%$ and $4.8\%$ of the communication and computation requirements, respectively, of the full state homogeneous fusion. 

\begin{figure}[bt]
	\centering
	\includegraphics[width=0.48\textwidth]{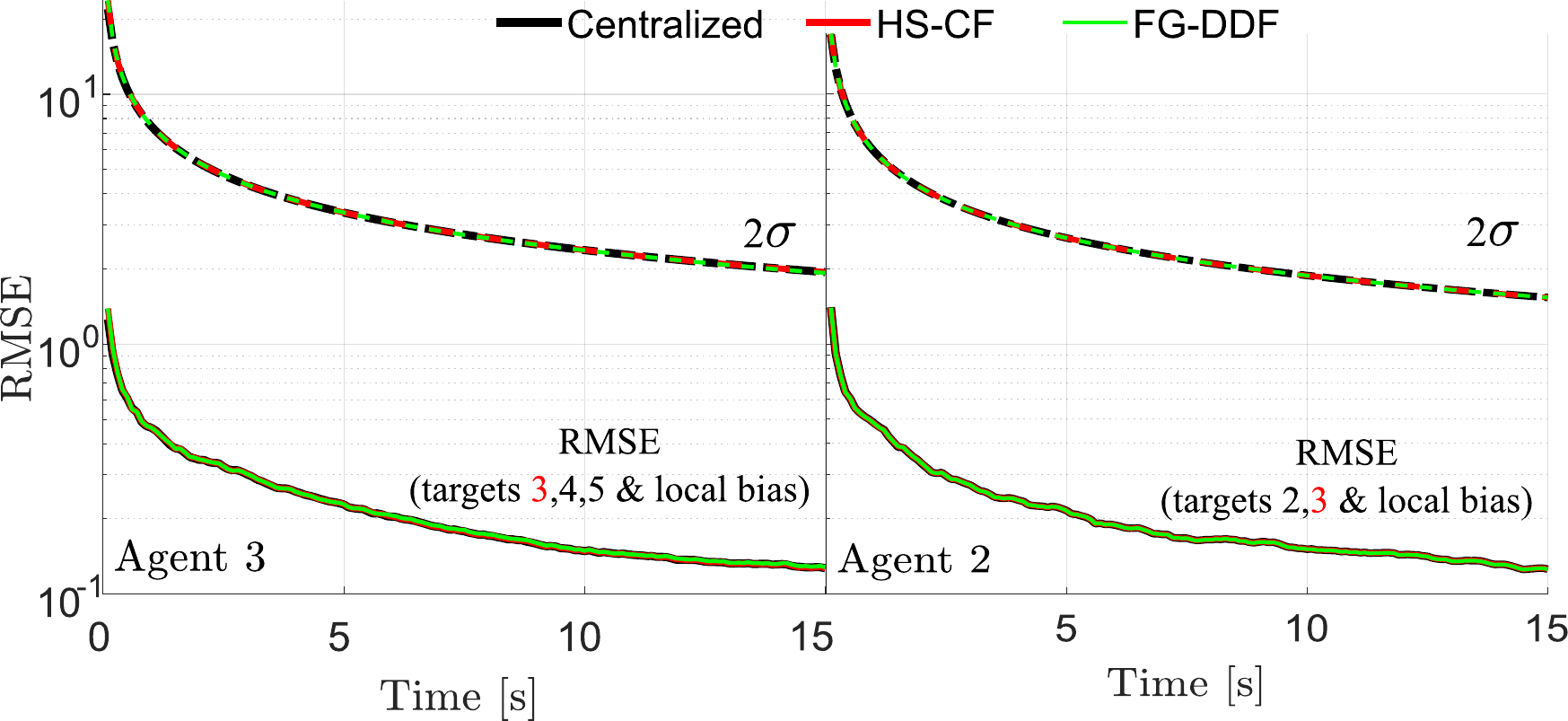}
	\caption{Validation of the FG-DDF algorithm. Shown are the RMSE results and 2$\sigma$ bounds of agents 2 and 3 over their local variables (with only target 3 in common), compared to a centralized estimate and the HS-CF algorithm from \cite{dagan_exact_2021}. } 
	\label{fig:staticSim}
	\vspace{-0.25in}
\end{figure}

Fig. \ref{fig:staticSim} validates our implementation by comparing the results of 50 Monte-Carlo simulations with the results from \cite{dagan_exact_2021}. We show that the mean RMSE (full line) and the $2\sigma$ lines of our new code FG-DDF, which has all its operations defined and performed on a factor graph, exactly matches the results of the HS-CF algorithm. Also plotted is a comparison with the centralized solution, showing the estimate is conservative, as it produces estimation error covariances which are larger in the positive semi-definite (psd) sense than the centralized (optimal) error covariance.

\begin{table}[bt]
\caption{Local platform target assignments and sensor measurement error covariances, taken from \cite{dagan_exact_2021}. }
    \begin{center}
    \begin{tabular}{c|c|c|c}
        Agent    & Tracked Targets & $R_i^{1} [m^2]$ & $R_i^{2} [m^2]$  \\ \hline
        1 & $T_1,T_2$ & diag([1,10]) & diag([3,3]) \\ \hline
        2 & $T_2,T_3$ & diag([3,3]) & diag([3,3]) \\ \hline
        3 & $T_3,T_4, T_5$ & diag([4,4]) & diag([2,2]) \\ \hline
        4 & $T_4,T_5$ & diag([10,1]) & diag([4,4]) \\ \hline
        5 & $T_5,T_6$ & diag([2,2]) & diag([5,5]) \\ \hline
    \end{tabular}
    \end{center}
    \label{tab:measurementError}
    \vspace{-0.22in}
\end{table} 

%\begin{figure*}[tb]
%	\centering
%	\includegraphics[width=0.8\textwidth]{Figures/A5_100MC_RMSE_wBDF_CF.eps}
%	\caption{RMSE - Agents 5}
%	\label{fig:simResults5}
%\end{figure*}

\subsection{Example 2 - Cooperative Mapping}
\label{sec:mapping}
\begin{figure}[bt]
	\centering
	\includegraphics[width=0.47\textwidth]{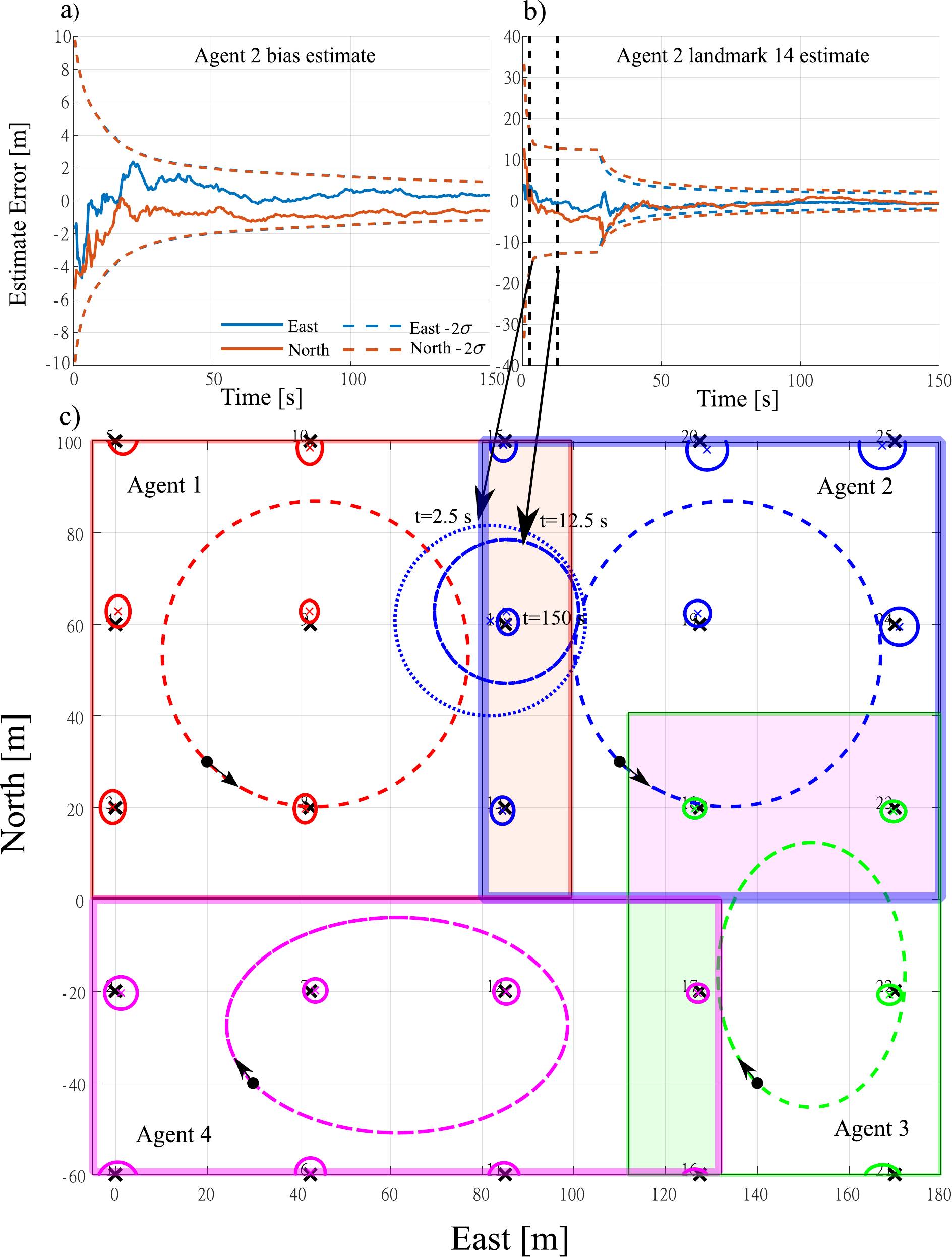}
	\caption{Multi-agent cooperative mapping illustration and simulation results. Agent 2 estimate and $2\sigma$ covariance bounds of a) landmark 14, and b) local bias. c) Global map and task subdivision with shaded overlapping sections. Black x's mark true landmark position and colored x's mark agent's estimate (at $t=150s$) with $2\sigma$ covariance ellipses.} 
	\label{fig:Mapping_results}
	\vspace{-0.28in}
\end{figure}

We test our algorithm with a simulation of a 2D multi-agent cooperative mapping problem. In this scenario, four agents are moving in a circular/elliptical trajectory in a  $180 m\times 160 m$ field. The field is subdivided into four separate but overlapping sections and each agent is tasked with mapping one of them (Fig. \ref{fig:Mapping_results}a). Here mapping refers to the task of inferring the $2D$ position of a subset of 25 known landmarks, scattered across the field.    
We assume each agent has perfect self position information, but with constant agent-landmark relative position measurement bias in the east and north direction. Further, we assume that an agent takes a noisy relative measurement only if the landmark is within $45m$ radius from the agent and that the data association problem is solved. In addition, each agent takes a noisy measurement to estimate its own bias.

In homogeneous DDF, each agent would need to reason about the full set of 25 landmarks, regardless of its relative distance and relevance to the local task as well as the other agents local bias state, resulting in 58 variables. Those variables then need to be processed and communicated repeatedly, incurring high overhead cost. In the heterogeneous FG-DDF approach, each agent only processes the locally relevant landmarks and biases (maximum 20) and communicates only information over the common variables (maximum 6), which in turns lead to about $96\%$ and $99\%$ computation and communication reduction, respectively.

Fig. \ref{fig:Mapping_results}a illustrates agent's trajectories and landmarks estimates after 150s with their $2\sigma$ confidence ellipses. Each agent only estimates its local map (subset of 6-9 landmarks). Fig. \ref{fig:Mapping_results}b and \ref{fig:Mapping_results}c show agent 2 estimate vs. time of its local bias and landmark 14, respectively. It is worth exploring the dynamics of the landmark's estimate of this case as we can observe three regions of the covariance bounds: 1) in the first 5s the agent is taking relative measurements to the landmark, allowing it to reduce its uncertainty to about 13.5m (two standard deviation). 2) In the time frame 5s-27s the agent has no direct new information on the landmark as it is out of measurement range, but it can nevertheless marginally improve its estimate to about 12.5m due to bias measurements and its correlation to the landmark state. 3) Agent 1, which has landmark 14 in common with agent 2, starts taking relative measurements to the landmark, agent 2 is then able to further improve its estimate by fusion with agent 1 to about 3m at $t=85$. This cycle repeats as the two agents get in and out of measurement range to landmark 14, but the changes are small.

\section{Conclusions}
\label{sec:conclusions}
In heterogeneous robotic systems, heterogeneous decentralized data fusion is key to enable data sharing and scalability for multi-agent real world task oriented applications such as cooperative mapping, target tracking and more. In this paper we merge the Bayesian decentralized data fusion (DDF) approach, which sees each agent as an independent entity, inferring the variable of interest based on its local information, with probabilistic graphical models, which frequently assumes knowledge of the full system graph and correlations. We show that factor graphs are a natural representation for heterogeneous fusion in robotic applications as it gives insight into the problem structure. This enables exploitation of conditional independence between sets of local and common variables of interest. We then suggest a shift in paradigm by slicing the global factor graph into smaller local graphs representing its individual, though cooperative, task. 

We assume linear Gaussian models and constant tree structured networks to develop the theory for factor graph based DDF (FG-DDF), including basic filtering, fusion and inference operations, performed solely on the graph.
We show how, by using this framework, fusion of information from a neighboring agent reduces into simply integrating a new factor into the local graph. We validate our algorithm by comparing to previous results of a static multi-agent multi-target simulation and further demonstrate its applicability with a cooperative mapping problem. In both cases, the results are consistent and conservative ((i.e. produce estimation error covariances which are larger in psd sense than the true error covariances)) and the advantages of heterogeneous fusion in terms of computational and communication scalability are highlighted.

Our approach enables further development and analysis of new fusion algorithms in more complicated scenarios, based on this relatively simple and general representation.
While we used some simplifying assumptions in order to develop the theory of FG-DDF, it is not limited to Gaussian pdf nor to tree-structured networks (e.g. covariance intersection could be used in place of the CF here \cite{loefgren_scalable_2019}). These avenues, as well as dynamic problems are the subject of future research.

%It requires agents to communicate and process only relevant data, and minimizing the amount of resources invested in variable less or inconsequential to the local agent's task.  

\bibliographystyle{IEEEtran}
\bibliography{references.bib}

\end{document}